\documentclass[final]{cvpr}

\usepackage{epsfig}
\usepackage{graphicx}
\usepackage{amsmath}
\usepackage{amssymb}

\usepackage{subcaption}
\usepackage{multirow}
\usepackage{ulem}
\usepackage{bm}

\usepackage[margin=3pt,font=small,labelfont=bf,labelsep=endash,tableposition=top]{caption}

\usepackage{xcolor}
\usepackage[colorlinks = true,
            linkcolor = purple,
            urlcolor  = blue,
            citecolor = cyan,
            anchorcolor = black]{hyperref}

\def\cvprPaperID{3651} % *** Enter the CVPR Paper ID here
\def\confYear{CVPR 2021}

\begin{document}

%%%%%%%%% TITLE
\title{Feature Decomposition and Reconstruction Learning for \\
Effective Facial Expression Recognition}

\author{Delian Ruan$ ^{1,2} $,
 ~~
Yan Yan$ ^{1} $\thanks{Corresponding author (email: {\tt yanyan@xmu.edu.cn}).},
 ~~
 Shenqi Lai$ ^ 2$,
  ~~
  Zhenhua Chai$ ^2 $,
   ~~
   Chunhua Shen$ ^3 $,
    ~~
    Hanzi Wang$ ^1 $\\
$^{1}$Xiamen University, China
~~~~$^{2}$Vision Intelligence Center, Meituan
~~~~
%, China\\
$^{3}$%The
University of Adelaide%, Australia
\\
%{\tt\small delianruan@stu.xmu.edu.cn, \{yanyan,wang.hanzi\}@xmu.edu.cn}\\
%{\tt\small\{laishenqi,chaizhenhua\}@meituan.com, chunhua.shen@adelaide.edu.au}
%\and
}

\maketitle
\pagestyle{empty}
\thispagestyle{empty}

%%%%%%%%% ABSTRACT
\begin{abstract}
   In this paper, we propose a novel Feature Decomposition and Reconstruction Learning (FDRL) method for effective facial expression recognition. We view the expression information as the combination of the shared information (expression similarities) across different expressions and the unique information (expression-specific variations) for each expression.
More specifically, FDRL mainly consists of two crucial networks: a Feature Decomposition Network (FDN) and a Feature Reconstruction Network (FRN).
In particular, FDN first decomposes the basic
features extracted from a backbone network into a set of facial action-aware latent features to model expression similarities.
Then, FRN captures the intra-feature and inter-feature relationships for  latent features to characterize expression-specific variations, and reconstructs the expression feature.
To this end, two modules including an intra-feature relation modeling module and an inter-feature relation modeling module are developed in FRN.
Experimental results on both the in-the-lab databases (including CK+, MMI, and Oulu-CASIA) and the in-the-wild
databases (including RAF-DB and SFEW) show that the
proposed FDRL method consistently achieves higher recognition accuracy than several state-of-the-art methods.
This clearly highlights the benefit of feature decomposition
and reconstruction for classifying expressions.
%
%\textcolor{red}{To this end, an Intra-feature Relation Modeling module  and an Inter-feature Relation Modeling module are developed in FRN, respectively.
%Experimental results on both the in-the-lab databases (including CK+, MMI, and Oulu-CASIA) and the in-the-wild databases (including RAF-DB and SFEW) show the superiority of our proposed method.
%%This clearly highlights the benefit of feature decomposition and reconstruction for FER.
\end{abstract}

%%%%%%%%% BODY TEXT
\section{Introduction}
{Facial expression is one of the most natural and universal signals for human beings to express their inner states and intentions  \cite{darwin1998expression}.
Over the past few decades, Facial Expression Recognition (FER) has received much attention in computer vision, due to its %widespread
various
applications including virtual reality, intelligent tutoring systems, health-care, \textit{etc.} \cite{zhang2018joint}. According to psychological studies \cite{ekman1971constants}, the FER task is to classify an input facial image into one of the following seven categories: angry (AN), disgust (DI), fear (FE), happy (HA), sad (SA), surprise (SU), and neutral (NE).}

\begin{figure}[t!]
\centering
\includegraphics[width=0.488\textwidth]{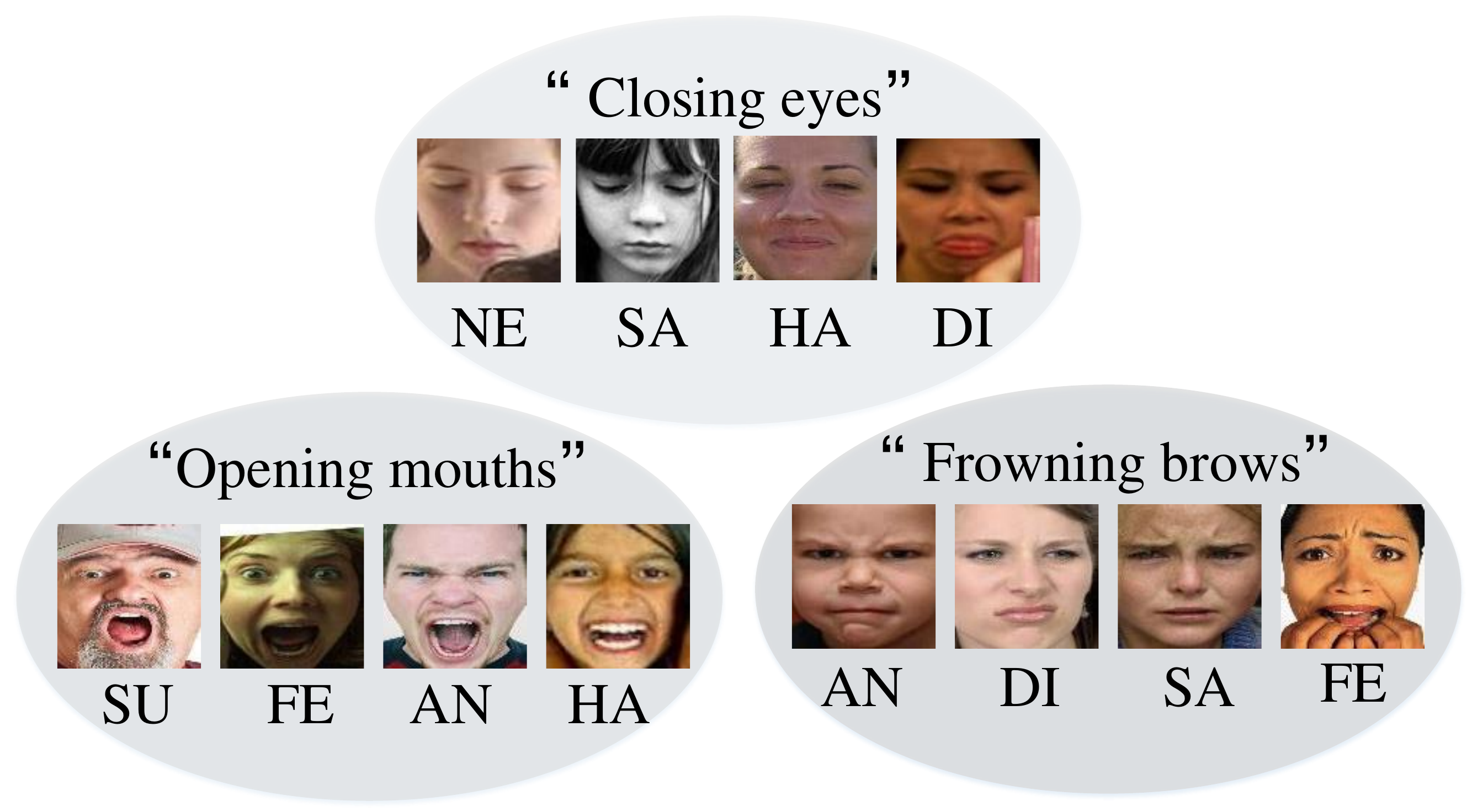}
\caption{The images in each group show a similar facial action, but they are from different expressions. Images are from the RAF-DB database \cite{li2018DLP-CNN}. }
\label{fig:challenges}
\end{figure}

A variety of FER methods \cite{cai2018island,li2018DLP-CNN,ruan2020DDL,xie2019DAM-CNN} have been proposed to learn holistic expression features by disentangling the disturbance caused by various disturbing factors, such as pose, identity, illumination, and so on. However, these methods %ignore
neglect
the fact that the extracted expression features corresponding to some expressions may still not be easily distinguishable, mainly because of high similarities across different expressions.

An example is %illustrated
shown
in Figure \ref{fig:challenges}. We can observe that some facial images corresponding to the NE, SA, HA, and DI expressions exhibit closing eyes. The facial images corresponding to the SU, FE, AN, and HA expressions all show opening mouths, while those corresponding to the AN, DI, SA, and FE expressions show frowning brows. The images from different facial expressions in each group give a similar facial action, where the distinctions between some expressions are subtle.
Therefore, how to learn effective fine-grained expression features to identify subtle differences in expressions by considering similar facial actions is of great importance.

The expression information is composed of the shared information (expression similarities) across different expressions and the unique information (expression-specific variations) for each expression.
The expression similarities can be characterized by shared latent features between different expressions, while the expression-specific variations can be reflected by {importance weights for latent features}. Therefore, the expression features can be represented by combining a set of latent features associated with their corresponding importance weights. Traditional FER methods  \cite{luo2013pca,mohammadi2014pca,zhao2008pca,deng2005lda} adopt Principal Component Analysis (PCA) or Linear Discriminant Analysis (LDA) to extract eigenvectors (corresponding to latent features) and eigenvalues (corresponding to importance weights).
However, these eigenvectors only capture holistic structural information rather than fine-grained semantic information of facial images, which is critical for FER.

Motivated by the success of deep learning in various vision tasks, %in this paper,
here we propose a novel Feature Decomposition and Reconstruction Learning (FDRL) method for effective FER. FDRL is mainly comprised of two crucial networks, including a Feature Decomposition Network (FDN) and a Feature Reconstruction Network (FRN). The two networks are tightly combined and jointly trained in an end-to-end manner.

Specifically,
a backbone convolutional neural network is first used to extract %the
basic features. Then, FDN decomposes the basic feature into a set of facial action-aware latent features, which effectively encode expression similarities across different expressions. In particular, a compactness loss is developed to obtain compact latent feature representations.
Next, FRN, which includes an Intra-feature Relation Modeling module (Intra-RM) and an Inter-feature Relation Modeling module (Inter-RM), models expression-specific variations and reconstructs the expression feature. Finally, an expression prediction network is %adopted
employed
for expression classification.

In summary, our  main contributions are summarized as follows.%of this paper are:}

\begin{itemize}\itemsep -0.1cm
\item {
%We formulate the FER problem from the perspective of signal decomposition, where the expression features are represented by a set of latent features associated with their corresponding importance weights. To achieve this,
 A novel FDRL method is proposed to perform FER. In FDRL, FDN and FRN are respectively developed to explicitly model expression similarities and expression-specific variations, enabling the extraction of fine-grained expression features.
 %In this way,
 Thus,
 the subtle differences between facial expressions can be accurately identified.}

\item{Intra-RM and Inter-RM are elaborately designed to learn an intra-feature relation weight and an inter-feature relation weight for each latent feature, respectively. Therefore, the intra-feature and inter-feature relationships between latent features are effectively captured to obtain discriminative expression features.}

\item {Our FDRL method is extensively evaluated on both the in-the-lab and the in-the-wild FER databases. Experimental results show that our method consistently outperforms several state-of-the-art FER methods. In particular, FDRL achieves 89.47\% and 62.16\% recognition accuracy on the RAF-DB and SFEW databases, respectively. This convincingly shows the great potentials of feature decomposition and reconstruction for FER.}
\end{itemize}

\section{Related work}
With the rapid development of deep learning, extensive efforts have been made to perform FER. State-of-the-art deep learning-based FER methods mainly focus on two aspects:
1) disturbance disentangling, and 2) expression feature extraction.

\subsection{{Disturbance Disentangling}}

{Many FER methods have been proposed to predict expressions by disentangling the disturbance caused by various disturbing factors, such as pose, identity, illumination, and so on.
Wang \textit{et al.}~\cite{wang2019IPFR} propose an adversarial feature learning method to
%handle
tackle
the disturbance caused by facial identity and pose variations. %Meng and Liu \cite{meng2017IACNN} propose an Identity-Aware Convolutional Neural Network (IACNN) to alleviate the influence of facial identity, where an identity-sensitive contrastive loss and a cross-entropy loss are adopted to supervise the model training.
Ruan \textit{et al.}~\cite{ruan2020DDL} propose a novel Disturbance-Disentangled Learning  (DDL)
method to simultaneously disentangle multiple disturbing factors. Note that the above methods depend largely on the label information of disturbing factors.} A few methods address the occlusion problem of FER. Wang and Peng \cite{wang2020RAN} propose a novel Region Attention Network  (RAN)
to adaptively adjust the importance of facial regions to mitigate the problems of occlusion and variant poses for FER.
 %Li \textit{et al.}~\cite{li2018gACNN} propose a convolution neutral network with attention mechanism method to identify the occlusion regions and extract features from the most discriminative non-occluded regions. }

{Recently, some methods are concerned with the noisy label problem in the FER databases.
Zeng \textit{et al.}~\cite{zeng2018IPA2LT} propose an Inconsistent Pseudo Annotations to Latent Truth  (IPA2LT)
method to deal with the problem of inconsistency in different FER databases.
Wang  \textit{et al.}~\cite{wang2020SCN} introduce a Self-Cure Network (SCN)
 to prevent the trained model from over-fitting uncertain facial images.
%Chen \textit{et al.}~\cite{chen2020LDL-ALSG} develop a novel Label Distribution Learning on Auxiliary Label Space Graphs (LDL-ALSG) method that exploits the topological information of the labels from auxiliary tasks. }

{The above methods perform FER by alleviating the influence caused by disturbing factors or noisy labels. However, they %might ignore
do not take into account
 subtle differences between different facial expressions.
In this paper, we formulate the FER problem from the perspective of feature decomposition and reconstruction, which successfully models expression similarities and expression-specific variations. Therefore, high-level semantic information can be effectively encoded to classify facial expressions.}

\begin{figure*}[th!]
\centering
\includegraphics[width=0.9\textwidth]{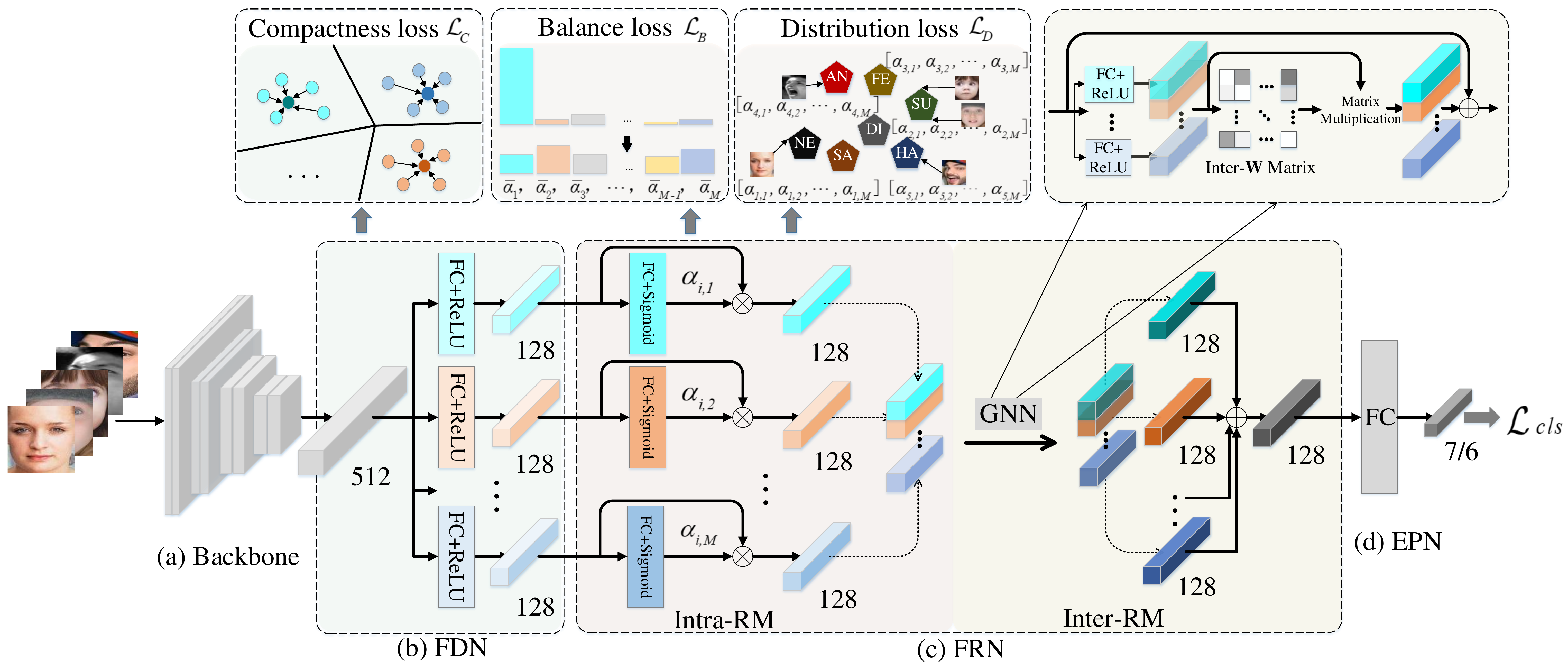}
\caption{Overview of our proposed FDRL method. (a) The backbone network (ResNet-18) that extracts
basic CNN features; (b) A Feature Decomposition Network (FDN) that decomposes the basic feature into a set of facial action-aware latent features; (c) A Feature Reconstruction Network (FRN) that learns an intra-feature relation weight and an inter-feature relation weight for each latent feature, and reconstructs the expression feature. FRN contains two modules:  an Intra-feature Relation Modeling module (Intra-RM) and an Inter-feature Relation Modeling module (Inter-RM); (d) An Expression Prediction Network (EPN) that predicts an expression label.
}
\label{fig:architecture}
\end{figure*}

\subsection{Expression Feature Extraction}

Some FER methods design effective network architectures and loss functions to reduce inter-class similarities and enhance intra-class compactness for expression feature extraction.
Li \textit{et al.}~\cite{li2018DLP-CNN} propose a deep locality-preserving loss based method, which extracts discriminative expression features by preserving the locality closeness.
%Huang \textit{et al.}~\cite{huang2018expression} develop a joint loss which combines the center loss and a novel inter-class loss to explicitly reduce intra-class variations and enlarge inter-class differences.
Cai \textit{et al.}~\cite{cai2018island} design a novel island loss to simultaneously increase inter-class separability and intra-class compactness.

A few FER methods employ attention mechanisms \cite{xie2019DAM-CNN} to improve the discriminative ability of  expression features. Xie \textit{et al.}~\cite{xie2019DAM-CNN} design an attention layer to focus on  salient regions of a facial expression.  Wang \textit{et al.}~\cite{wang2020RAN} determine the importance of different facial regions by leveraging an attention network.

The above methods enhance the discriminative capability of expression features by designing different loss functions or attention mechanisms. These methods  consider the expression features as holistic features.
In contrast,
%motivated by the geometrical structure of facial expression images,
we decompose the basic features into a set of facial action-aware latent features and then model the intra-feature and inter-feature relationships for latent features. Compared with holistic features used in traditional methods, the latent feature representations developed in our method are more fine-grained and facial action-aware. Such a manner is beneficial to learn expression features for identifying subtle differences between facial expressions.
%(coarse-grained features)

\section{Our Method}
% CS: THIS DOESN"T CONVERY MSG
% We first provide an overview of our proposed FDRL method and then describe the details of key components in our method.

% \subsection{Overview}
\noindent {\bf Overview.}
{The proposed FDRL method consists of a backbone network, a Feature Decomposition Network (FDN), a Feature Reconstruction Network (FRN), and an Expression Prediction Network (EPN). An overview of the proposed method is shown in Figure \ref{fig:architecture}.}

Given a batch of facial images, we first feed them into a backbone network (in this paper, we use ResNet-18 \cite{he2016resnet} as the backbone) to extract basic CNN features.  Then, FDN decomposes the basic features into a set of facial action-aware latent features, where a compactness loss is designed to extract compact feature representations. Next, FRN learns an intra-feature relation weight and an inter-feature relation weight for each latent feature, and reconstructs the expression feature. Finally, EPN (a simple linear fully-connected layer) predicts a facial expression label.

In particular, FRN consists of two modules: an Intra-feature Relation Modeling module (Intra-RM) and an Inter-feature Relation Modeling module (Inter-RM). To be specific, Intra-RM is first introduced to assign an intra-feature relation weight to each latent feature according to the importance of the feature, and thus an intra-aware feature is obtained. To ensure similar distributions of intra-feature relation weights for facial images from the same expression category, a distribution loss and a balance loss are employed in Intra-RM. Then, Inter-RM computes an inter-feature relation weight by investigating the relationship between intra-aware features, and thus an inter-aware feature is extracted. At last, the expression feature is represented by a combination of the intra-aware feature and the inter-aware feature. FRN exploits both the contribution of each latent feature and the correlations between intra-aware features, enabling the extraction of discriminative expression features.

\subsection{{Feature Decomposition Network (FDN)}}
Given the $i$-th facial image, the basic feature %obtained
extracted
by the backbone network is denoted as $\textbf{x}_i \in \mathbf{R}^{P}$, where $P$ is the dimension of the basic feature. As %we
mentioned previously, FDN decomposes the basic feature into a set of facial action-aware latent features. Let $\textbf{L}_i=[\textbf{l}_{i,1}, \textbf{l}_{i,2}, \cdots, \textbf{l}_{i,M}] \in \mathbf{R}^{D \times M}$ denote a facial action-aware latent feature matrix, where $\textbf{l}_{i,j}\in \mathbf{R}^D$ represents the $j$-th latent feature for the $i$-th facial image. $D$ and $M$ represent  the dimension of each latent feature and the number of latent features, respectively.

Specifically, to extract the $j$-th latent feature, we employ a linear Fully-Connected (FC) layer and a ReLU activation function, which can be formulated as:
\begin{equation}
\textbf{l}_{i,j} = \sigma_1(\mathbf{W}_{{d}_{j}}^\textrm{T} \textbf{x}_i) \quad \textrm{for} \ j = 1, 2, \cdots, M,
\label{eq:destruct}
\end{equation}
where $\mathbf{W}_{{d}_{j}}$ denotes the parameters of the FC layer used for extracting the $j$-th latent feature and $\sigma_1$ represents the ReLU function. %All the latent features are extracted in the same way.

\noindent {\textbf{Compactness Loss.} Since different facial expressions share the same set of latent features, it is expected that a set of compact latent feature representations %is
are
extracted. In other words, the $j$-th latent feature extracted from one basic feature should be similar to that extracted from another basic feature. To achieve this, inspired by the center loss \cite{wen2016center}, we develop a compactness loss. The compactness loss $\mathcal{L}_C$ learns a center for the same latent features and penalizes the distances between the latent features and their corresponding centers, which can be formulated as:
\begin{equation}
\mathcal{L}_C = \frac{1}{N}\sum_{i=1}^{N}\sum_{j=1}^{M}\parallel \textbf{l}_{i,j} - \textbf{c}_{j}\parallel_2^2,
\end{equation}
where $N$ denotes the number of images in a mini-batch. $\parallel \cdot \parallel_2$ indicates the $L_2$ norm. $\mathbf{c}_{j} \in \mathbf{R}^D$ denotes the center of the $j$-th latent features, and is updated based on a mini-batch.
% In this way,
Thus,
the intra-latent variations are minimized and a set of compact latent features are effectively learned.}

%We visualize the latent features on an input image from RAF-DB, as shown in Fig.
%It is clearly that different latent features focus on different regions of a facial image. For example, the first several latent features are concerned with the eyes and eyebrows, while some latent features pay attention to the mouths and noses. Therefore, the latent features correspond to fine-grained and facial action-aware features, which can be useful for classifying facial expressions.
{To visually demonstrate the interpretation of latent features, we collect a group of images that corresponds to the highest intra-feature relation weight (see Section \ref{sub:FRN}) of the same latent feature and then visualize them. In Figure \ref{fig:groups}, we can observe that the images from each group show a specific facial action. The images from the nine groups show the facial actions of ``Neutral'', ``Lip Corner Puller'', ``Staring'', ``Opening Mouths'', ``Lips Part'', ``Closing Eyes'', ``Grinning'', ``Frowning Brows'', and ``Lip Corner Depressor'', respectively.}
%For instance,  all the images from group 1 show frowning brows.  The images from group 2 exhibit opening mouths. The images from group 3 give the facial action of staring and those from group 4 contain the facial action of grinning.
Therefore, the latent features obtained by FDN are fine-grained and facial action-aware features, which can be useful for subsequent expression feature extraction.

\begin{figure}[t!]
\centering
\includegraphics[width=0.48\textwidth]{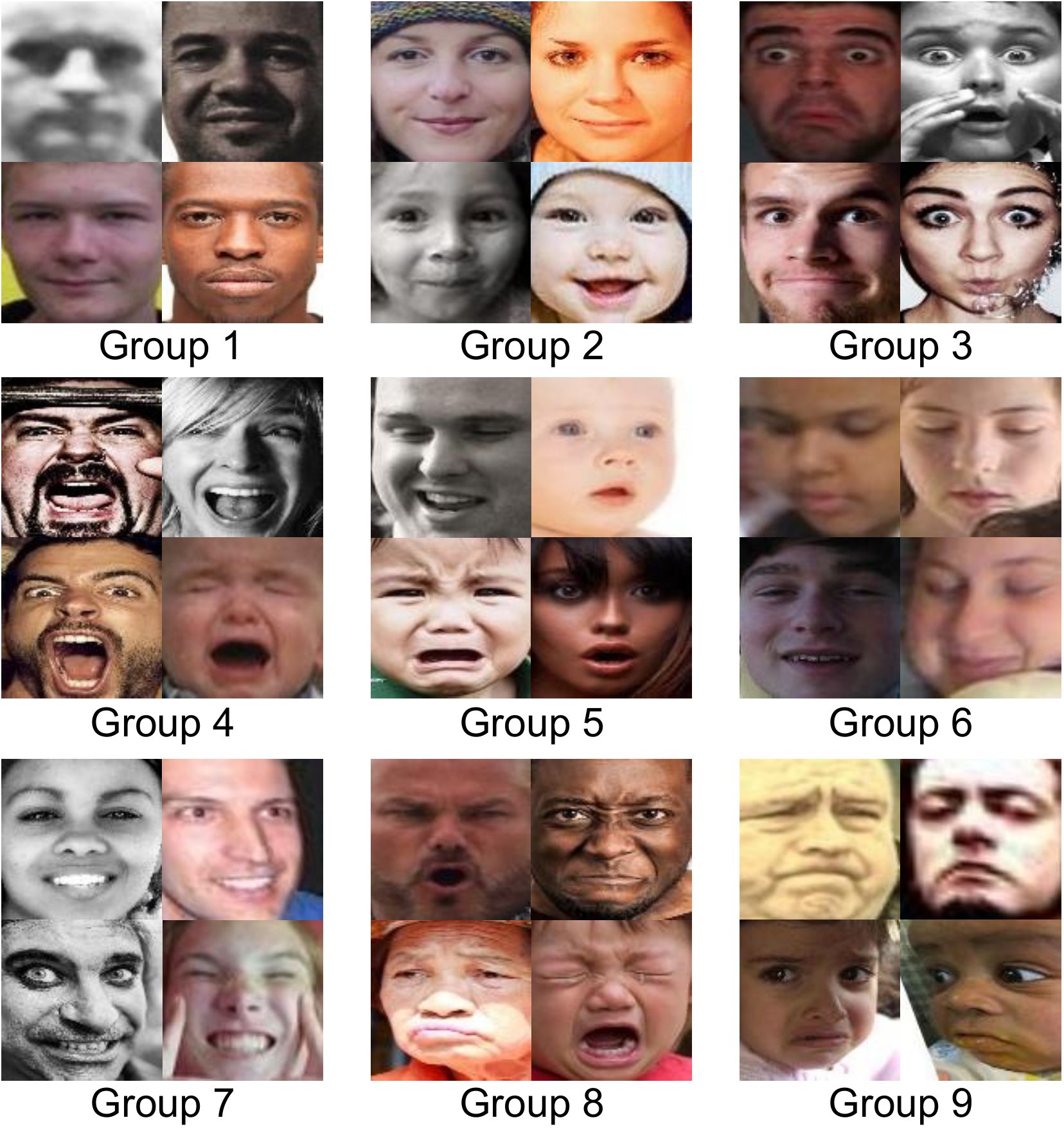}
\caption{{Visualization of the image groups from the RAF-DB database when $M$ is set to 9. Each group corresponds to the highest intra-feature relation weight of the same latent feature.}}
\label{fig:groups}
\end{figure}

\subsection{{Feature Reconstruction Network (FRN)}}
\label{sub:FRN}
In this section, FRN, which models expression-specific variations, is %elaborately
carefully
designed to obtain discriminative expression features. FRN contains two modules: Intra-RM and Inter-RM.

\noindent \textbf{{Intra-feature Relation Modeling Module (Intra-RM).}}
{Intra-RM consists of multiple intra-feature relation modeling blocks, where each block is designed to model the intra-feature relationship between feature elements.

To be specific, each block is composed of an FC layer and a sigmoid activation function, that is:
\begin{equation}
\bm{\alpha}_{i,j}= \sigma_2(\mathbf{W}_{s_{j}}^\textrm{T} \textbf{l}_{i,j}) \quad \textrm{for} \ j = 1, 2, \cdots, M,
\label{eq:3}
\end{equation}
where $\bm{\alpha}_{i,j} \in \mathbf{R}^{D}$ denotes the importance weights for the $j$-th latent feature corresponding to the $i$-th facial image, $\mathbf{W}_{s_{j}}$ represents the parameters of the FC layer, and $\sigma_2$ indicates the sigmoid function.}

% Based on
With
Eq.~(\ref{eq:3}), we compute the $L_1$ norm of $\bm{\alpha}_{i,j}$ as the Intra-feature relation Weight (Intra-W) to determine the importance of the $j$-th latent feature, that is:
\begin{equation}
\alpha_{i,j} = \parallel \bm{\alpha}_{i,j}\parallel_1 \quad \textrm{for} \ j = 1, 2, \cdots, M,
\end{equation}
where $\parallel \cdot \parallel_1$ denotes the $L_1$ norm.

It is desirable that the distributions of Intra-Ws corresponding to different images from the same expression category are as close as possible. Therefore, similarly to the compactness loss, a distribution loss is used to learn a center for each expression category and penalize the distances between the Intra-Ws from one class and the corresponding center. Hence, the variations caused by different disturbing factors are alleviated.

Suppose that
the $i$-th facial image belongs to the {$k_i$}-th expression category. Mathematically, the distribution loss $\mathcal{L}_D$ is formulated as:
\begin{equation}
\mathcal{L}_D = \frac{1}{N}\sum_{i=1}^{N}\parallel\mathbf{w}_i-\mathbf{w}_{{k_{i}}}\parallel_2^2,
\end{equation}
where $\textbf{w}_i=[\alpha_{i,1}, \alpha_{i,2}, \cdots, \alpha_{i,M}]^{\textrm{T}} \in \mathbf{R}^{M}$ represents the Intra-W vector for the $i$-th facial image. {$\mathbf{w}_{{k_{i}}}\in \mathbf{R}^{M}$ denotes the class center corresponding to the {$k_i$}-th expression category.}%; $K$ is the number of expression categories.

By optimizing the distribution loss, the Intra-W vectors corresponding to different images from the same expression category are closely distributed. Thus,  they are able to focus on expression-specific variations.

In practice, as shown in Figure \ref{fig:architecture}, some Intra-Ws (corresponding to one or two latent features) usually show much higher values than the other Intra-Ws in the Intra-W vector for each image, since these Intra-Ws are individually computed. To alleviate this problem, we further design
a balance loss to balance the distributions of elements in each Intra-W vector as:
\begin{equation}
\mathcal{L}_B = \parallel \overline{\textbf{w}}- \textbf{w}_{u}\parallel_1,
\end{equation}
{where $\overline{\textbf{w}}= [\overline{\alpha}_1,\overline{\alpha}_2, \cdots, \overline{\alpha}_M]^{\textrm{T}}\in \mathbf{R}^{M}$ represents the mean Intra-W vector for a batch of samples (i.e., $\overline{\textbf{w}}=\frac{1}{N}\sum_{i=1}^{N}\mathbf{w}_i)$.}  $\textbf{w}_u=[\frac{1}{M}, \frac{1}{M}, \cdots, \frac{1}{M}]^{\textrm{T}}\in \mathbf{R}^{M}$ denotes a uniformly-distributed weight vector.

{After computing an Intra-W for each latent feature, we assign this weight to the corresponding feature and obtain an intra-aware feature for the $i$-th facial image as:
\begin{equation}
\textbf{f}_{i,j} = \alpha_{i,j} \textbf{l}_{i,j} \quad \textrm{for} \ j = 1, 2, \cdots, M,
\end{equation}
where $\textbf{f}_{i,j} \in \mathbf{R}^{D}$ represents the $j$-th intra-aware feature for the $i$-th facial image.}

\noindent \textbf{{Inter-feature Relation Modeling Module (Inter-RM).}}
Intra-RM learns an Intra-W for each individual latent feature. However, these Intra-Ws are independently extracted. Although the distribution loss imposes consistency regularization on the Intra-W, it does not fully consider the inter-relationship between latent features. In fact, for each facial expression, different kinds of facial actions usually simultaneously appear. For example,
 the FE expression often involves the facial actions of frowning brows and opening mouths. The HA expression contains the facial actions of stretching brows, closing eyes, and opening mouths. Therefore, it is critical to exploit the correlations between different facial action-aware latent features. To achieve this,  we further introduce  Inter-RM to learn an Inter-feature Relation Weight (Inter-W) between intra-aware features based on Graph Neural Network (GNN) \cite{bruna2013gnn,shen2018gnn}.
%
%
%
%For FE expressions¡ê?"frowning brows" and "opening mouths" usually appear .  And for SU expressions, "staring eyes" and "opening mouths" usually appear simultaneouly.
%
%
%\textcolor{red}{For example, for the expression of HA, the latent feature corresponding to ``opening mouth'' may be assigned with a high intra-feature relation weight and that corresponding to ``closing eyes'' may be assigned with a low intra-feature relation weight.
%However, the facial actions of opening mouth and closing eyes can simultaneously appear for some facial images with the expression of HA (see ``closing eyes'' of HA in Figure \ref{fig:challenges}). , so that the latent features corresponding to the opening mouth and closing eyes can be properly activated for extracting expression features, while those corresponding to other facial actions are suppressed.}

%, as illustrated in Figure \ref{fig:architecture2}.}

%For example, the "staring" often appears with "opening mouth", which means that if the "staring" latent feature has a high weight importance, the "opening mouth" part should also be assigned equivalent weight importance.

%Specifically, given the primary discriminative features $F^1 = [f_0^1, f_1^1, \cdots, f_n^1]$, we regard the $F^1$ as the set of nodes and construct an undirected complete graph $G(F^1, E)$, where E denotes the edges between all the nodes.

{Inter-RM learns a set of relation messages and estimates the Inter-Ws between these messages. Specifically, for each $\textbf{f}_{i,j}$, it is first fed into a message network for feature encoding. In this paper, the message network is composed of an FC layer and a ReLU activation function, which is:
\begin{equation}
{\textbf{g}}_{i,j}= \sigma_1(\mathbf{W}_{e_{j}}^\textrm{T} \textbf{f}_{i,j}) \quad \textrm{for} \ j = 1, 2, \cdots, M,
\end{equation}
where $\mathbf{W}_{e_{j}}$ denotes the parameters of the FC layer used for feature encoding and $\sigma_1$ represents the ReLU function. ${\textbf{g}}_{i,j} \in \mathbf{R}^{D}$ denotes the $j$-th relation message for the $i$-th facial image.}

Then, a relation message matrix ${\textbf{G}}_{i}  =  [{\textbf{g}}_{i,1}, {\textbf{g}}_{i,2},\cdots,$ $ {\textbf{g}}_{i,M}]\in \mathbf{R}^{D\times M}$ is represented as nodes in the graph $G({\textbf{G}}_{i}, E)$. In our formulation, $G$ is an undirected complete graph and $E$ represents the set of relationships between different relation messages. $\omega_i(j,m)$ is the Inter-W which denotes the relation importance between the node ${\textbf{g}}_{i,j}$ and the node ${\textbf{g}}_{i,m}$. It can be calculated as:
\begin{equation}
\omega_i(j,m) =
\begin{cases}
\sigma_3(S({\textbf{g}}_{i,j}, {\textbf{g}}_{i,m})) & j\not=m \\
0 &  j=m
\end{cases},
\label{eq:w}
\end{equation}
where ${\textbf{g}}_{i,j}$ and ${\textbf{g}}_{i,m}$ are the $j$-th and the $m$-th relation messages for the $i$-th facial image, respectively. $S$ is a distance function, which estimates the similarity score between ${\textbf{g}}_{i,j}$ and ${\textbf{g}}_{i,m}$. In our paper, we use the Euclidean distance function. Since the results of $S(\cdot)$ are all positive, we further adopt the tanh activation function $\sigma_3$ to normalize the positive distance value to [0,1). The purpose of setting $\omega_i(j,j)$ to 0 is to avoid self-enhancing. According to Eq.~(\ref{eq:w}), an Inter-W matrix $\textbf{W}_i = \{\omega_i(j,m)\} \in  \mathbf{R}^{M \times M}$ can be obtained.

Hence, the $j$-th inter-aware feature $\hat{\textbf{f}}_{i,j}\in \mathbf{R}^{D}$ for the $i$-th facial image can be formulated as:
\begin{equation}
\hat{\textbf{f}}_{i,j} = \sum_{m=1}^M\omega_i(j,m)\textbf{g}_{i,m} \quad \textrm{for} \ j = 1, 2, \cdots, M.
\end{equation}

By combining the $j$-th intra-aware feature and the $j$-th inter-aware feature, the $j$-th importance-aware feature $\textbf{y}_{i,j}\in \mathbf{R}^{D}$ for the $i$-th facial image is calculated as:
\begin{equation}
\textbf{y}_{i,j}= \delta \textbf{f}_{i,j} + (1-\delta)\hat{\textbf{f}}_{i,j} \quad \textrm{for} \ j = 1, 2, \cdots, M,
\label{eq:gnn}
\end{equation}
where $\delta$ represents the regularization parameter that balances the intra-aware feature and the inter-aware feature.}

{Finally, a set of importance-aware features are added to obtain the final expression feature, that is,
\begin{equation}
\textbf{y}_i = \sum_{j=1}^M\textbf{y}_{i,j},
\end{equation}
where $\textbf{y}_i\in \mathbf{R}^{D}$ represents the expression feature for the $i$-th facial image.}

\subsection{Joint Loss Function}
In the proposed FDRL, the backbone network, FDN, FRN, and EPN are jointly trained in an end-to-end manner.
The whole network minimizes the following joint loss function:
\begin{equation}
\mathcal{L}  = \mathcal{L}_{cls} + \lambda_1\mathcal{L}_C + \lambda_2\mathcal{L}_B + \lambda_3\mathcal{L}_D,
\label{total loss}
\end{equation}
where $\mathcal{L}_{cls}$,  $\mathcal{L}_C$, $\mathcal{L}_B$, and $\mathcal{L}_D$ represent the classification loss, the compactness  loss, the balance loss, and the distribution loss, respectively. {In this paper, we use the cross-entropy loss as the classification loss.} $\lambda_1$, $\lambda_2$,  and  $\lambda_3$ denote the regularization parameters. By optimizing the joint loss, FDRL is able to extract discriminative fine-grained expression features for FER.

\section{Experiments}
We first briefly introduce five public FER databases.
%, including three in-the-lab databases (CK+, MMI, and Oulu-CASIA) and two in-the-wild databases (RAF-DB and SFEW).
Then, we describe the implementation details,
%Next, we
and
perform ablation studies with qualitative and quantitative results to show the importance of each component in FDRL.
Finally, we compare FDRL with state-of-the-art FER methods.

\subsection{Databases}
{\noindent \textbf{CK+ \cite{lucey2010ck}} contains 327 video sequences, which are captured in controlled lab environments.
%The CK+ database is provided with seven expression categories, including six basic expressions (i.e., angry, happy, surprise, sad, disgust, and fear) and one non-basic expression (i.e., contempt).
We choose the three peak expression frames from each expression sequence to construct the training set and the test set, thus resulting in a total of 981 images.
\noindent \textbf{MMI \cite{pantic2005mmi}} is also a lab-controlled database, containing 205 video sequences with six basic expressions. We choose the three peak frames from each sequence to construct the training set and the test set,  thus resulting in a total of 615 images.
%The subject-independent ten-fold cross-validation is also used.
\noindent \textbf{Oulu-CASIA \cite{zhao2011oulu}} contains videos captured in controlled lab conditions.
%with the three illumination conditions using two imaging systems.
We select the last three frames in each sequence captured with the visible light and strong illumination to construct the training set and the test set (consisting of 1,440 images in total).
% The subject-independent ten-fold cross-validation is conducted in our experiments.
Similarly to \cite{ding2017facenet2expnet,meng2017IACNN,yang2018DeRL,zhao2016PPDN}, we employ the subject-independent ten-fold cross-validation protocol for evaluation on all the three in-the-lab databases.}
\begin{table*}[!t]
\centering
\small
\caption{Ablation studies for the different values of $\lambda_1$, $\lambda_2$, and $\lambda_3$ (represent the balance parameters for compactness loss, balance loss, and distribution loss, respectively) on MMI and  RAF-DB. The recognition accuracy (\%) is used for performance evaluation.}\label{tab:ablation}
\begin{subtable}[!t]{.3\textwidth}
\centering
		\setlength{\abovecaptionskip}{0.1cm}
		{
				\centering
				 \caption{Influence of $\lambda_1$.}\label{tab:similar}
				 \renewcommand\arraystretch{1.06}
				 \setlength{\tabcolsep}{2mm}
{
\begin{tabular}{c|cc}
\hline
$\lambda_1$          & MMI  		&	  RAF-DB						\\
\hline
\hline
$0$		&   84.64  &     88.75        	\\
$0.00001$		&     85.02   &    89.02     	\\
$0.0001$		&  \textbf{85.23}   &     \textbf{89.47}	       	\\
$0.001$		&   82.67 &     88.82	         	\\
$0.01$ 	&   82.24   &     88.63	       	\\
\hline
\end{tabular}
}}
\end{subtable}\begin{subtable}[!t]{.3\textwidth}
\centering
		\setlength{\abovecaptionskip}{0.1cm}
		{
				\centering
				 \caption{Influence of $ \lambda_2$.}\label{tab:balance}
				 \renewcommand\arraystretch{1.06}
				 \setlength{\tabcolsep}{3mm}
{
\begin{tabular}{c|cc}
\hline
$\lambda_2$         & MMI  	  &	 RAF-DB 							\\
\hline
\hline
0					&		82.66	&		88.23			\\
0.5				&   83.68  &    88.89	       	\\
1.0				&  \textbf{85.23} &		\textbf{89.47}	         	\\
1.5			    &   84.94  	&    88.92 	        	\\
2.0      		&   83.23  	&    88.63	     	\\
\hline
\end{tabular}}}
\end{subtable}\begin{subtable}[!t]{.3\textwidth}
\centering
		\setlength{\abovecaptionskip}{0.1cm}
		{
				\centering
				 \caption{Influence of $\lambda_3$.}\label{tab:consistency}
				 \renewcommand\arraystretch{1.06}
				 \setlength{\tabcolsep}{2mm}
{
\begin{tabular}{c|cc}
\hline
$\lambda_3$            & MMI  	&	  RAF-DB							\\
\hline
\hline
$0$		&   84.96   &     89.15	       	\\
$0.00001$		&   85.07&     88.89          	\\
$0.0001$		&  \textbf{85.23}  &     \textbf{89.47}	        	\\
$0.001$		&   82.66   &     88.95       	\\
$0.01$	&   81.64   &     88.49	       	\\
\hline
\end{tabular}}}
\end{subtable}
\end{table*}

{\noindent \textbf{RAF-DB \cite{li2018DLP-CNN}} is a real-world FER database, which contains 30,000 images labeled with basic or compound expressions by 40 trained human labelers. The images with six basic expressions and one neutral expression are used in our experiment. RAF-DB involves 12,271 images for training and 3,068 images for testing.
\noindent \textbf{SFEW \cite{dhall2011SFEW}} is created by selecting static frames from Acted Facial Expressions in the Wild (AFEW) \cite{dhall2012AFEW}.
%SFEW is a challenging FER database, where the images are captured by varied head poses, large age range, real-world illumination, and so on.
The images in SFEW are labeled with six basic expressions and one neutral expression. We use 958 images for training and 436 images for testing.}

\subsection{Implementation Details}
For each database, all the facial images are detected and cropped according to eye positions, and the cropped images are further resized to the size of $256 \times 256$. During the training process, the facial images are randomly cropped to the size of $224 \times 224$, and then a random horizontal flip is applied for data augmentation. During the test process, the input image is center cropped to the size of $224 \times 224$ and then fed into the trained model.
The FDRL method is implemented with the Pytorch toolbox and the backbone network is a lightweight ResNet-18 model \cite{he2016resnet}. {Similarly to \cite{wang2020SCN}, the ResNet-18 is pre-trained on the MS-Celeb-1M face recognition database \cite{guo2016ms}.}

%The values of $\lambda_1$ and $\lambda_2$ in Eq.\ref{total loss} are both empirically set to 0.0001.
The dimension of the basic feature is 512. The dimensions of both the latent feature and the expression feature are 128. The value of $\delta$ in Eq. (\ref{eq:gnn}) is empirically set to 0.5.
 We train our FDRL in an end-to-end manner with a single TITAN X GPU for 40 epochs, and the batch size for all the databases is set to 64. The model is trained using the Adam algorithm \cite{kingma2014adam} with the initial learning rate of 0.0001, $\beta_1=0.500$, and $\beta_2=0.999$.  The learning rate is further divided by 10 after 10, 18, 25, and 32 epochs.

\subsection{Ablation Studies}
To show the effectiveness of our method, we perform ablation studies to evaluate the influence of key parameters and components on the final performance. For all the experiments, we use one in-the-lab database (MMI) and one in-the-wild database (RAF-DB) to evaluate the performance.
%{The values of $\lambda_1$ and $\lambda_2$ represent the importance of the compact loss and the consistency loss, respectively.}}

\begin{figure}[t!]
\centering
\includegraphics[scale=0.6]{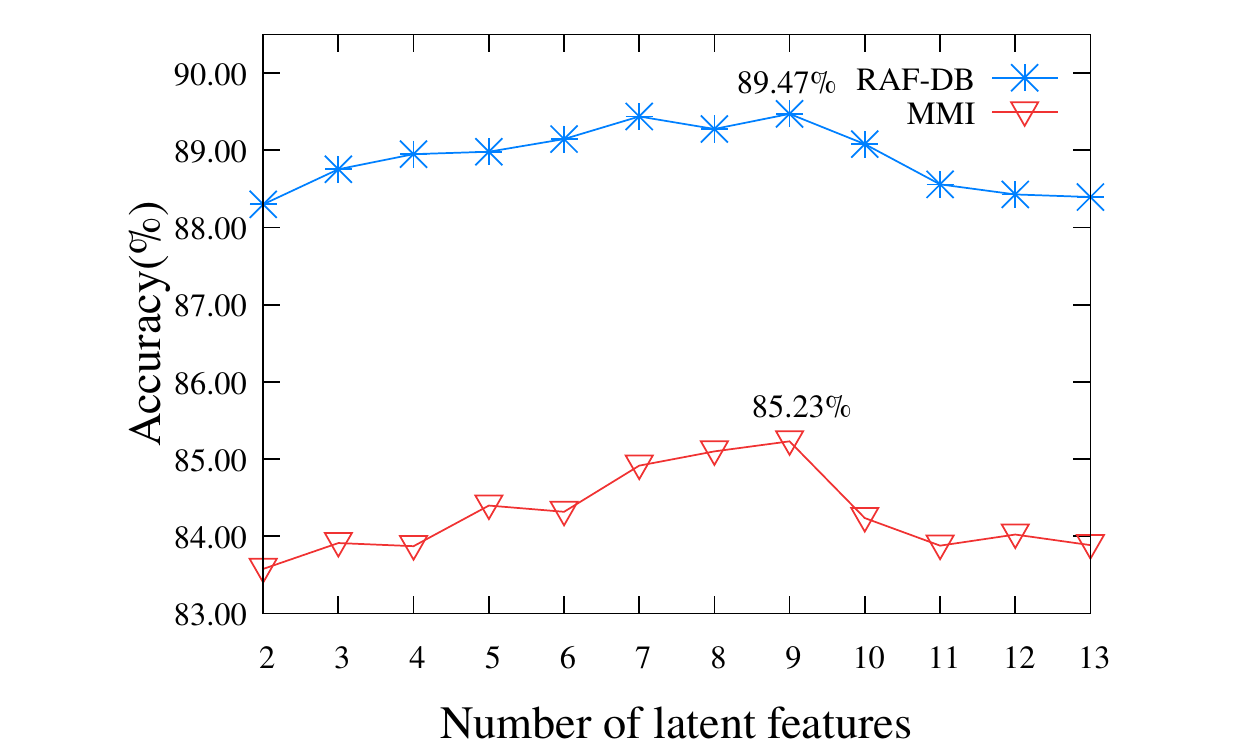}
\caption{Ablation studies for the different numbers of latent features on the MMI and RAF-DB databases.}
\label{fig:num_feats}
\end{figure}

\noindent \textbf{Influence of the number of latent features.}
%We evaluate the recognition accuracy of the proposed method with the different numbers of latent features,
As shown in Figure \ref{fig:num_feats}, we can see that the proposed method achieves the best recognition accuracy when the number of latent features is set to 9.
On  one hand, when a small number of latent features are used, the expression similarities cannot be effectively modeled. On the other hand, when a large number of latent features are used, there exist redundancy and noise among latent features, thus leading to a performance decrease. In the following experiments, we set the number of latent features to 9.

\noindent \textbf{Influence of the parameters.}
We evaluate the recognition performance of the proposed method with the different values of $\lambda_1$, $\lambda_2$, and $\lambda_3$ in Eq.~(\ref{total loss}), as shown in Table \ref{tab:ablation}.
% $\lambda_1$ and $\lambda_2$ represent the regularization parameters of the compact loss and the distribution loss, respectively. $\delta$ denotes the regularization parameter that balances the intra-aware feature and the inter-aware feature.

Specifically, we first fix $\lambda_2=1.0$ and $\lambda_3 = 0.0001$, and set the value of $\lambda_1$ from 0 to 0.01. Experimental results are given in Table \ref{tab:ablation} (\subref{tab:similar}). We can observe that our method achieves the best performance when the value of $\lambda_1$ is set to 0.0001. When $\lambda_1=0$, our method is trained without using the compactness loss, and the performance decreases.  Table \ref{tab:ablation} (\subref{tab:balance}) shows the recognition performance obtained by our method, when the values of $\lambda_1$ and $\lambda_3$ are both set to 0.0001, and the value of $\lambda_2$ varies from 0 to 2.0. When the value of $\lambda_2$ is set to 1.0, our method achieves the top performance. Then, we fix $\lambda_1 = 0.0001$ and $\lambda_2=1.0$, and set the value of $\lambda_3$ from 0 to 0.01. Experimental results are given in Table \ref{tab:ablation} (\subref{tab:consistency}). Our method obtains the top performance when $\lambda_3=0.0001$.
In the following, we set the values of both $\lambda_1$ and $\lambda_3$ to 0.0001, and set the value of $\lambda_2$ to 1.0.

%As shown in Table \ref{tab:ablation} (\subref{tab:similar}), we evaluate $lambda_1$ from 0 to 0.01 on the models whose number of latent features is set to 9 and the value of $lambda_2$ is set to 0.0001. $lambda_1=0$ means that we train the model without similar loss. Experimental results show that when the value of $\lambda_1$ is set to 0.0001, the model achieves the best performance. It is clear that training the model with $\lambda_1$ greater than 0.01 is not a good choice, which leading to worse performance than the model without similar loss.

\begin{table}[!t]
\centering
\small
\caption{Ablation studies for three key modules of our  FDRL on the MMI and RAF-DB databases. The recognition accuracy (\%) is used for performance evaluation.}\label{tab:ablation_module} %The baseline model is ResNet18.
\centering
		\setlength{\abovecaptionskip}{0.1cm}
		{
				 \renewcommand\arraystretch{1.06}
				 \setlength{\tabcolsep}{3mm}
{
\begin{tabular}{ccc|cc}
\hline
\multirow{2}{*}{FDN} &  \multicolumn{2}{c|}{FRN} & 	\multirow{2}{*}{MMI}  &	 \multirow{2}{*}{RAF-DB}	\\
	&Intra-RM		&	Inter-RM         &   		 &	 					\\
\hline
\hline
$\times$	&$\times$	&		$\times$	&		79.69	&		86.93		\\
$\surd$		&$\times$	&		$\times$	&		81.23	&		87.71		\\
$\surd$		&$\times$	& $\surd$			&  	83.44  	&     88.76	        	\\
$\surd$		&$\surd$	& 	$\times$	&   	84.74   	&    89.34       	\\
$\surd$		&$\surd$	& $\surd$      	&  \textbf{85.23} &    \textbf{89.47}         	\\
\hline
\end{tabular}}}
\end{table}

\begin{table*}[!t]
\centering
\small
\caption{Performance comparisons among different methods on several public FER databases. The best results are boldfaced. $\ddagger$ and $\dagger$ respectively denote that seven expression categories and six expression categories are used in CK+.
 %* indicates that the method is trained based on the image sequences.
 }
 \label{tab:sota}
\small
		\begin{subtable}[!t]{.5\textwidth}
		\setlength{\abovecaptionskip}{0.1cm}
				{
				\centering
				 \caption{Comparisons on the in-the-lab databases.}\label{tab:lab}
				 \renewcommand\arraystretch{1.06}
\begin{tabular}{c|ccc}
\hline
\multirow{2}{*}{Methods}            &	 \multicolumn{3}{c}{Accuracy (\%)}		 	\\
  																& CK+ 	& MMI 	& Oulu-CASIA 								\\
\hline
\hline
PPDN \cite{zhao2016PPDN}        			&     97.30$\dagger$				&    	- 					&   72.40        	\\
IACNN \cite{meng2017IACNN}             	&     95.37$\ddagger$			&   71.55			&     -     				\\
DLP-CNN \cite{li2018DLP-CNN}				&		95.78$\dagger$				&		78.46		&		-						\\
%DTAGN* \cite{jung2015DTAGN}					&		97.25$\ddagger$				&		70.20		&		81.46			\\
IPA2LT \cite{zeng2018IPA2LT}					&		92.45$\ddagger$				&   65.61			&    61.49    		\\
DeRL \cite{yang2018DeRL}       				&    	 97.37$\ddagger$			&   73.23  		&     88.00      	\\
%PHRNN-MSCNN* \cite{Zhang2017PHRNN-MSCNN}		&98.50$\ddagger$	&81.18		&86.25		\\
FN2EN \cite{ding2017facenet2expnet}              &     98.60$\dagger$				&    - 					&    87.71     		\\
DDL \cite{ruan2020DDL}													&		99.16$\ddagger$ 			&		83.67	&	\textbf{88.26}		\\
\hline
\hline
Baseline															&		97.15$\ddagger$							&			79.69			&		86.18										\\
FDRL (proposed)													&		\textbf{99.54}	$\ddagger$ 	&		\textbf{85.23}		&		\textbf{88.26}		\\
\hline
\end{tabular}}
\end{subtable}\quad
\begin{subtable}[!t]{.4\textwidth}
\setlength{\abovecaptionskip}{0.1cm}
\centering
\caption{Comparisons on the in-the-wild databases.}\label{tab:wild}
\setlength{\tabcolsep}{5mm}
{
\renewcommand\arraystretch{1.06}
\begin{tabular}{c|cc}
\hline
\multirow{2}{*}{Methods}            &	 \multicolumn{2}{c}{Accuracy (\%)}	 \\
														& RAF-DB 	& SFEW			\\
\hline
\hline
IACNN \cite{meng2017IACNN}				&		-					&			50.98								\\
DLP-CNN \cite{li2018DLP-CNN}				&		84.13		&			51.05								\\
%gACNN \cite{li2018gACNN}						&		85.07		&			-											\\
IPA2LT \cite{zeng2018IPA2LT}        		&   86.77  		&     	58.29		     					\\
SPDNet \cite{acharya2018SPDNet} 	&  		87.00 			&   58.14	     	\\
RAN \cite{wang2020RAN}							&		86.90		&			56.40									\\
SCN \cite{wang2020SCN}                                            &     87.01      &        -                                       \\
DDL \cite{ruan2020DDL}							       		       &    87.71	 	&        59.86                              \\
\hline
\hline
Baseline																	&		86.93		&			58.03									\\
FDRL (proposed)													&		\textbf{89.47}		&			\textbf{62.16}		\\
\hline
\end{tabular}}
\end{subtable}
\end{table*}

\noindent \textbf{{Influence of the key modules}.}
%We verify the effectiveness of the Intra-RM module and the Inter-RM module in FRN.
%We evaluate the four variants of our proposed method:
%(1)  the FDRL method without using both Intra-RM and Inter-RM in FRN (denoted as FDRL\_w/o\_Intra\&Inter), which directly combines the original latent features in FDN as the final expression feature;
%(2) the FDRL method without using Intra-RM in FRN (denoted as FDRL\_w/o\_Intra);
%(3) the FDRL method without using Inter-RM in FRN (denoted as FDRL\_w/o\_Inter); and
%(4) the proposed FDRL method.
To evaluate the effectiveness of the key modules in FDRL, we perform ablation studies for FDN, Intra-RM, and Inter-RM on the MMI and RAF-DB databases, respectively. Experimental results are reported  in Table \ref{tab:ablation_module}.

We can see that incorporating FDN into the backbone network improves the performance, which shows the importance of FDN.
Moreover, by employing  Intra-RM or Inter-RM in FRN, we are able to achieve better recognition accuracy than the method combining FDN and the backbone network. This is because the features extracted by FDN are not distinguishable enough to classify different expressions, since FDN does not take expression-specific variations into account.
In contrast, Intra-RM and Inter-RM effectively model the intra-feature relationship of each latent feature and the inter-feature relationship between intra-aware features, respectively, leading to performance improvements.
Our proposed FDRL method, which combines the backbone network, FDN, and FRN in an integrated network, achieves the best results among all the variants.

% We can see that the added of FDN (2nd row) into the baseline (1st row) can degrade performance slightly.
% %Besides, the added of Inter-RM and Intra-RM
% However, compared with the methods with Intra-RM or Inter-RM in FRN (3rd row and 4th row), the method which only add FDN obtains the worst performance. This is  }

%FDRL\_w/o\_Inter outperforms FDRL\_w/o\_Intra\&Inter by $3.51\%$ and $1.63\%$ improvements on the MMI and RAF-DB databases, respectively. This shows the importance of Intra-RM. Compared with FDRL\_w/o\_Intra\&Inter, FDRL\_w/o\_Intra improves the accuracy by  $2.21\%$ and $1.05\%$ on MMI and RAF-DB, respectively. In addition, FDRL outperforms FDRL\_w/o\_Inter  by $0.49\%$ and $0.13\%$ on the MMI and RAF-DB databases, respectively. Therefore, Inter-RM can adequately model the relationship between different intra-aware features. Among all the variants, FDRL\_w/o\_Intra\&Inter achieves the worst performance. This is because the expression features extracted by FDRL\_w/o\_Intra\&Inter are not distinguishable enough to classify different expressions, since it does not take expression-specific variations into account.
%Both FDRL\_w/o\_Inter and FDRL\_w/o\_Intra outperform FDRL\_w/o\_Intra\&Inter on two databases. This shows the importance of Intra-RM and Intra-RM. Among all the variants, FDRL\_w/o\_Intra\&Inter achieves the worst performance. This is because the expression features extracted by FDRL\_w/o\_Intra\&Inter are not distinguishable enough to classify different expressions, since it does not take expression-specific variations into account.}

\begin{figure}[t!]
\centering
\includegraphics[scale=0.35]{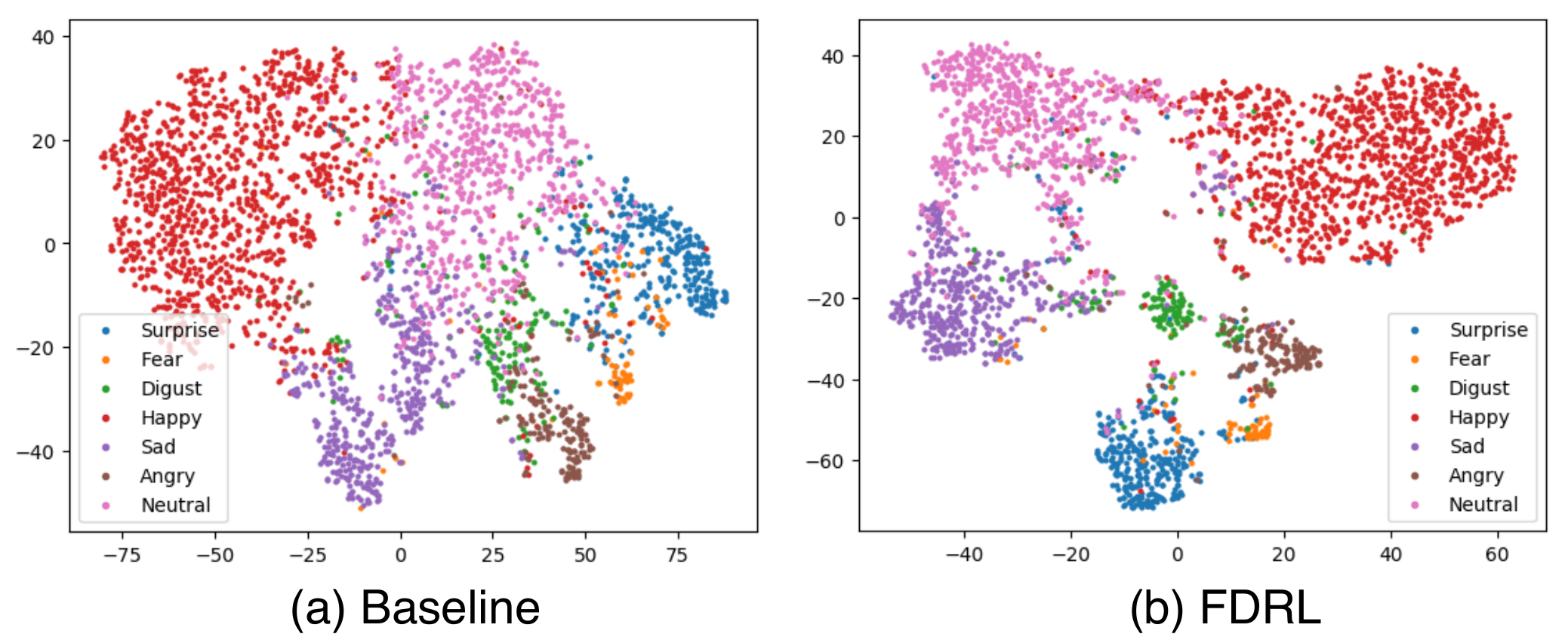}
\caption{Visualization of the expression features using t-SNE.
%The expression features are extracted by using the baseline method and the proposed FDRL method on the RAF-DB database, respectively.
Features are extracted from the RAF-DB database.}
\label{fig:feats}
\end{figure}

\begin{figure}[t!]
\centering
\includegraphics[scale=0.6]{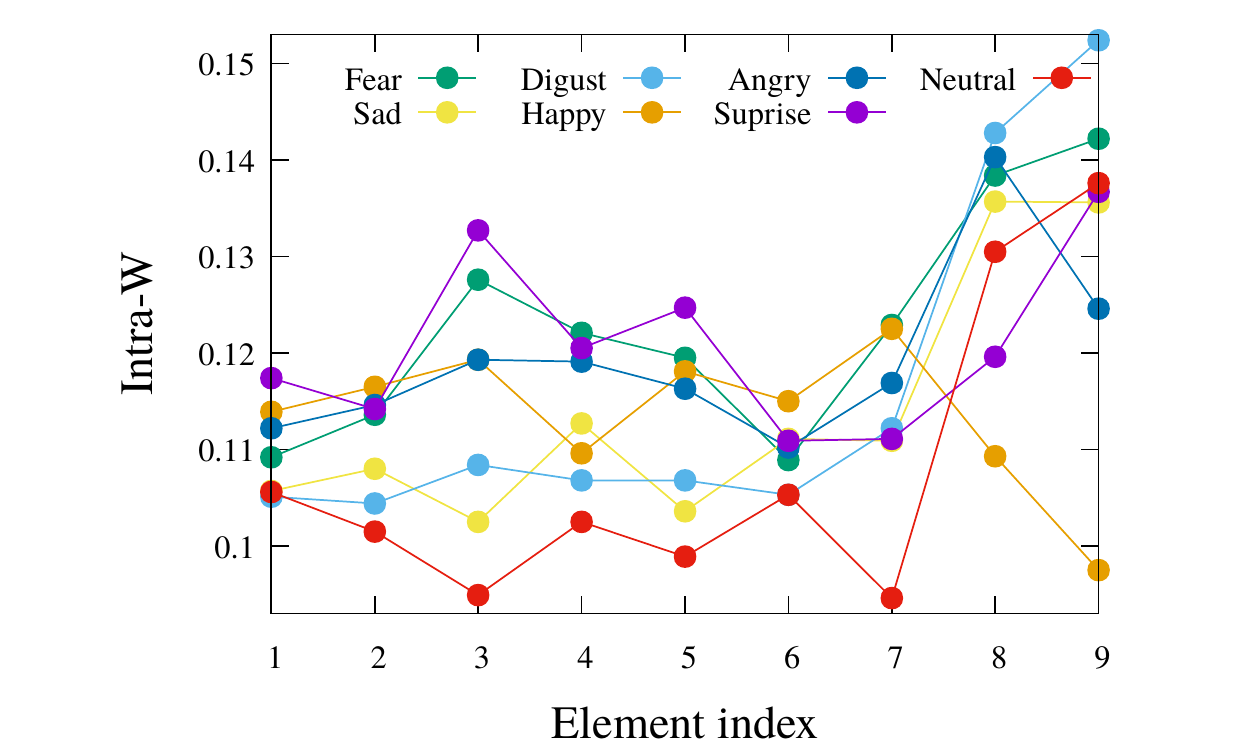}
\caption{{Visualization of distribution of mean Intra-W vectors for seven basic expression categories on the RAF-DB database.}}
\label{fig:weight_distri}
\end{figure}

\subsection{Visualization}
\noindent \textbf{2D feature visualization.} We use t-SNE \cite{maaten2008tsne} to visualize the expression features extracted by the baseline method (which only adopts ResNet-18) and the proposed FDRL method on the 2D space, respectively, as shown in Figure \ref{fig:feats}. We can observe that the expression features extracted from the baseline method are not easily distinguishable for different facial expressions. In contrast, the features extracted from our proposed method can effectively reduce intra-class differences and enhance inter-class separability for different expressions. Especially, compared with baseline, the differences between fear and surprise, disgust and sadness are more distinct for FDRL.

\noindent {\textbf{Distribution of mean Intra-W vectors.} We visualize the distribution of mean Intra-W vectors (corresponding to nine latent features) for seven basic expression categories on the RAF-DB database, as shown in Figure \ref{fig:weight_distri}. Generally, each expression shows relatively high weights on the latent features associated with facial actions (as shown in Figure \ref{fig:groups}) closely related to this expression. Nevertheless, we can observe that some latent features (such as 2nd and 6th, 1st and 4th) have similar weights for different expressions. Hence, we further develop  Inter-RM to exploit the inter-feature relationship between different intra-aware
features.}

%\noindent {\textbf{Interpretation of latent features.}

\subsection{{Comparison with State-of-the-Art Methods}}
{Table \ref{tab:sota} shows the comparison results between our method and several state-of-the-art FER methods on the in-the-lab databases and the  in-the-wild databases.}

{Among all the competing methods, IACNN, DDL, and RAN aim to disentangle the  disturbing factors in facial expression images. SCN and IPA2LT are proposed to solve the noise label problem.  FN2EN,  DTAGN, and  SPDNet improve the model performance by designing new network architectures.  DLP-CNN alleviates  intra-class variations by using a novel loss function. The above methods improve the FER performance by suppressing the influence of different disturbing factors or noise labels, but they ignore large expression similarities among different expressions. In contrast, our method explicitly models expression similarities and expression-specific variations with FDN and FRN, respectively, leading to performance improvements.}

{PPDN is developed to focus on the differences between expression images. DeRL claims that a facial expression is composed of the expression component and the neutral component. These two methods extract coarse-grained expression features. On the contrary, our proposed FDRL extracts more fine-grained features based on feature decomposition and reconstruction. Such a manner is beneficial to discriminate subtle differences between facial expressions, especially similar expression categories (such as fear and surprise). The above experimental results show the effectiveness of our proposed method.}

\section{Conclusion}
{In this paper, we have proposed a novel FDRL method for effective FER. FDRL consists of two main networks: FDN and FRN. FDN effectively models the shared information across different expressions based on a compactness loss. FRN accurately characterizes the unique information for each expression by taking advantage of Intra-RM and Inter-RM, and reconstructs the expression feature. In particular,
 Intra-RM encodes the intra-feature relationship of each latent feature and obtains an intra-aware feature. Inter-RM exploits the inter-feature relationship between different intra-aware features and extracts an inter-aware feature.
 The expression feature is represented by combining the intra-aware feature and the inter-aware feature.
 %By jointly training these networks in an end-to-end manner, our proposed method is able to extract discriminative expression features.
 Experimental results on both the in-the-lab and the in-the-wild databases have shown the superiority of our method to perform FER.}

%------------------------------------------------------------------------
\section*{Acknowledgements}
\small This work was in part supported by the National Natural Science Foundation of China under Grants 62071404 and 61872307, by the Natural Science Foundation of Fujian Province under Grant 2020J01001, by the Youth Innovation Foundation of Xiamen City under Grant 3502Z20206046, and by the Beijing Science and Technology Project under Grant Z181100008918018.

{%\small
\normalem
\bibliographystyle{ieee_fullname}
\bibliography{egbib}
}

\end{document}